\RequirePackage{fix-cm}
\documentclass[conference, twocolumn]{svjour3}          
\smartqed  
\usepackage{times}
\usepackage{epsfig}
\usepackage{graphicx}
\usepackage{amsmath}
\usepackage{amssymb}
\usepackage{enumitem}
\usepackage{float}
\usepackage{caption}
\usepackage{subcaption}
\usepackage{url}
\usepackage{cite}
\usepackage{multirow}
\usepackage{longtable}

\def\makeheadbox{\relax}

\begin{document}
\sloppy

\title{3D Human motion anticipation and classification}

\author{Emad Barsoum \and
        John Kender \and
        Zicheng Liu
}

\institute{Emad Barsoum \at
              One Microsoft Way, WA 98052 \\
              \email{ebarsoum@caa.columbia.edu}
           \and
           John Kender \at
              116th St \& Broadway, New York, NY 10027 \\
              \email{jrk@cs.columbia.edu}
           \and
           Zicheng Liu \at
              One Microsoft Way, WA 98052 \\
              \email{zliu@microsoft.com}
}

\date{Original 2019}

\maketitle

\begin{abstract}
Human motion prediction and understanding is a challenging problem. Due to the complex dynamic of human motion and the non-deterministic aspect of future prediction. We propose a novel sequence-to-sequence model for human motion prediction and feature learning, trained with a modified version of generative adversarial network, with a custom loss function that takes inspiration from human motion animation and can control the variation between multiple predicted motion from the same input poses. 

Our model learns to predict multiple future sequences of human poses from the same input sequence. We show that the discriminator learns general presentation of human motion by using the learned feature in action recognition task. Furthermore, to quantify the quality of the non-deterministic predictions, we simultaneously train a motion-quality-assessment network that learns the probability that a given sequence of poses is a real human motion or not.

We test our model on two of the largest human pose datasets: NTURGB-D and Human3.6M. We train on both single and multiple action types. Its predictive power for motion estimation is demonstrated by generating multiple plausible futures from the same input and show the effect of each of the loss functions. Furthermore, we show that it takes less than half the number of epochs to train an activity recognition network by using the feature learned from the discriminator.
\end{abstract}

\section{Introduction}
\label{intro}
Accurate short-term (several second) predictions of what will happen in the world given past events is a fundamental and useful human ability. Such aptitude is vital for daily activities, social interactions, and ultimately survival. For example, driving requires predicting other cars' and pedestrians' motions in order to avoid an accident; handshaking requires predicting the location of the other person's hand; and playing sports requires predicting other players' reactions. In order to create a machine that can interact seamlessly with the world, it needs a similar ability to understand the dynamics of the human world, and to predict probable futures based on learned history and the immediate present.

However, the future is not deterministic, so predicting the future cannot be deterministic, except in the very short term. As the predictions extend further into the future, uncertainty becomes higher. People walking may turn or fall; people throwing a ball may drop it instead.  However, some predictions are more plausible than others, and have a higher probability.

In this paper, we focus on creating a model that can predict multiple plausible future human (skeleton) poses from a given past. The number of poses taken from the immediate past, and the predicted number of poses in the future, which can be unrestricted, are parameters to the model. To accomplish this, we use a modified version of the generative adversarial network (GAN)\cite{nips2014:Goodfellow} and the improved Wasserstein generative adversarial network (WGAN-GP)\cite{corr2017:Ishaan} with a custom loss function that takes into consideration human motion and human anatomy.

The generator is a novel adaptation of sequence-to-sequence model\cite{nips2014:Ilya} of poses derived from a Recurrent Neural Network (RNN), and the critic and discriminator are a multilayer network (MLP). We use the critic network to train the generator and the discriminator network to learn distinguishing between a real sequence of poses from a fake one. In essence, we combine some aspect of the original GAN~\cite{nips2014:Goodfellow} with WGAN-GP\cite{corr2017:Ishaan}. We train our model on all actions at once, so its output is not conditioned on any specific human action. Our model takes as input a sequence of previous skeletal poses, plus a random vector $z$ from the reduced sequence space which samples possible future poses. For each such $z$ value, the model generates a different output sequence of possible future poses.

We use an RNN for the generator because RNNs are a class of neural networks designed to model sequences, especially variable length sequences. They have been successfully used in machine translation \cite{nips2014:Ilya}, caption generation from images \cite{iccv2015:Jia, cvpr2015:Donahue}, video classification and action recognition\cite{hbu2011:Baccouche, cvpr2015:Yue-Hei, cvpr2015:Donahue}, action detection\cite{cvpr2016:Yeung}, video description\cite{cvpr2016:Pan, corr2014:Kiros, corr2015:Yao, cvpr2015:Donahue}, sequence prediction\cite{corr2013:Graves, icml2015:Srivastava} and others.

We structure the learning by using a GAN because GANs\cite{nips2014:Goodfellow} are a class of unsupervised learning algorithms, inspired by game theory\cite{pjm1958:Sion}, which allow the generation of futures that are not tied to specific ground truth. Among other domains, they have been used to generate impressive realistic images.

However, GANs are difficult to train and unstable in their learning, their loss value does not necessarily indicate the quality of the generated sample, and the training can collapse easily. Recent literature\cite{corr2017:Arjovsky,corr2017:Ishaan,corr2017:Neyshabur,corr2016:Uehara,corr2017:Qi,corr2016:Mao} tries to improve GAN training and provide a theoretical guaranty for its convergence. In our work, we address this by adding a custom loss based on the skeleton physics in addition to the GAN loss, in order to stabilize and improve the training.

To quantitatively assess the quality of the non-deterministic predictions, we simultaneously train a motion-quality-assessment model that learns the probability that a given skeleton sequence is a real human motion.

We test our motion prediction model on two large datasets each captured with a different modality. The first is the NTURGB-D\cite{cvpr2016:Shahroudy} dataset, which is the largest available RGB-D and skeleton-based dataset, with data captured by using the Microsoft Kinect v2 sensor. The second is the Human3.6M\cite{pami2014:Ionescu,iccv2011:Ionescu} dataset, which is one of the largest available datasets derived from motion capture (MoCap) data.

Our main contributions are:
\setlist{nolistsep}
\begin{enumerate}[noitemsep]
\item We propose a novel human motion model that can predict multiple possible futures from a single past.
\item We propose a motion-quality-assessment model to quantitatively evaluate the quality of the predicted human motions.
\item We show that the feature learned by the discriminator can be applied to improve action recognition task.
\end{enumerate}

\section{Related work}
\label{sec:related_work}

Since the introduction of the Kinect sensor, there has been much work on recognizing human action and predicting human poses from skeleton data. For example, predicting human poses conditioned on previous poses using deep RNNs~\cite{iccv2015:Katerina, cvpr2016:Ashesh, cvpr2017:julieta} is due in part to this availability of large human motion datasets~\cite{pami2014:Ionescu,iccv2011:Ionescu,cvpr2016:Shahroudy}. In general, human motion prediction can be categorized into two categories: probabilistic and deterministic prediction.

\subsection{Probabilistic motion prediction}

Most work in probabilistic human motion prediction uses non-deep learning approaches~\cite{nips2000:Pavlovic, eccv2002:Sidenbladh, nips2005:Wang, pami2008:Wang, icml2013:Hema, cvpr2014:Lehrmann, iccv2015:Katerina}. In~\cite{cvpr2014:Lehrmann}, the authors propose simple Markov models that model observed data, and use the proposed model for action recognition and task completion. The limitation in this approach is that it predicts motion from a single action only, and assumes that human motion satisfies the Markov assumption. In~\cite{pami2008:Wang}, the authors introduce Gaussian process dynamical models (GPDMs) to model human pose and motion. However, they train their model on each action separately, and constrain the model to a Gaussian process. In~\cite{eccv2002:Sidenbladh}, the authors map human motion to a low dimensional space, and use the coordinates as an index into a binary tree that predicts the next pose, thus casting the prediction problem into a search problem. However, this approach can not generalize to previously unseen motions. In~\cite{nips2000:Pavlovic}, the authors use switching linear dynamic systems learned through a Bayesian network, and in~\cite{icml2013:Hema} the author used conditional random fields (CRF) to model spatio-temporal dynamics.

In contrast to all the above, our work does not use any statistical models to constrain the motion.

\subsection {Deterministic motion prediction}

Recent human motion prediction, which relies on deep RNNs~\cite{iccv2015:Katerina, cvpr2016:Ashesh, cvpr2017:julieta} or deep neutral networks~\cite{corr2017:Judith, corr2018:Li}, is primarily deterministic. In~\cite{iccv2015:Katerina}, the authors mix both deterministic and probabilistic human motion predictions. Their deterministic aspect is based on a modified RNN called Recurrent-Decoder (ERD) that adds fully connected layers before and after an LSTM~\cite{nc1997:Hochreiter} layer and minimizes an Euclidean loss. Their probabilistic aspect uses a Gaussian Mixture Model (GMM) with five mixture components and minimizes the GMM negative log-likelihood. For both aspects, they predict a single future human pose at a time. To predict more, they recurrently feed the single predicted pose back to the input. One drawback of this approach is error drifting, where the prediction error of the current pose will propagate into the next pose. In contrast, we predict multiple human poses at once thus avoiding error drifting. In addition, we do not impose any statistical model constrains like GMM over the motions.

In~\cite{cvpr2016:Ashesh}, the authors develop a general framework that converts a structure graph to an RNN, called a Structure-RNN (S-RNN). They test their framework on different problem sets including human motion prediction, and show that it outperforms the current state of the art. However, they need to design the structure graph manually and task-specifically. In~\cite{cvpr2017:julieta}, the authors examine recent deep RNN methods for human motion prediction, and show that they achieve start-of-the-art results with a simpler model by proposing three simple changes to RNN. In~\cite{corr2017:Judith}, the authors use an encoder-decoder network based on a feed-forward network, and compare the results of three different such architectures: symmetric, time-scale, and hierarchical.

However, the main issues of deterministic prediction of human motion are two-fold. The future is not deterministic, so the same previous poses could lead to multiple possible poses. And, using an $L_{2}$ norm can cause the model to average between two possible futures~\cite{corr2015:Mathieu}, resulting in blurred motions.

\subsection {Non-human motion prediction}

The prediction of multiple possible futures using RNNs has precedents. In \cite{corr2013:Graves}, the authors use an LSTM\cite{nc1997:Hochreiter} to generate text and handwriting from an input sequence. They generate one item at a time, by sampling the resultant probability. Then, they append the predicted item to the input sequence and remove its first item, and iterate. This creates a sequence of desired length, but the method may eventually create an input to the LSTM that does not contain any of the original input. In contrast, we pursue a method that trains our model to generate the entire desired output sequence of poses all at once.

The prediction of single or multiple possible futures using GANs also has precedents. In\cite{corr2015:Mathieu}, the authors trained a convolution model for both the generator and the discriminator in order to predict future frames. They improved the predictions by adding an image gradient difference loss to the adversary loss.  However, they again only predict a single possible future and, due to the use of the convolution network, the model can only predict a fixed length output. In contrast, we support variable length input and variable length output, and can also generate multiple possible futures from the same input.

In \cite{corr2017:Chen}, the authors predict or imagine multiple frames from a single image. They generate affine transformations between each frame, and apply them to the original input image to produce their prediction. Although this can imagine multiple futures from the same input image, a single image is not sufficient to capture the temporal dynamics of a scene. Furthermore, it makes the oversimplified assumption that the change between the images can be captured by using an affine transform.

\subsection {Human motion quality assessment}

Compared to the amount of work on motion editing and synthesis in computer graphics, the research on automatic motion quality evaluation has received little attention~\cite{tog:Ren2005,tog:Harrison2004,tog:Hodgins1998,tog:Reitsma2003}. The existing techniques were typically designed for special types of motions, or for motions obtained from software used to edit character motion~\cite{cmm:Wang2014}. A novel aspect of our motion quality assessment model is that it is trained simultaneously together with the motion prediction model.

\section{Our approach}

In human motion prediction problem, the system takes a sequence of human poses as input and predicts multiple valid future poses. Let $x=\{x_{1},x_{2},...,x_{m}\}$ be the input sequence of human poses and $y=\{y_{1},y_{2},...,y_{n}\}$ be the predicted sequence, where the concatenated sequence of $x$ and $y$ corresponds to a single activity, each pose can be presented with 3D joint locations or joint angles. Future is not deterministic, our goal is to be able to sample multiple possible future $y$ from the same input sequence $x$. By that we mean that the model learn an implicit condition probability of the future sequence conditioned on the input sequence $P(y|x)$.

\begin{figure*}[ht]
\centering
\includegraphics[width=\textwidth]{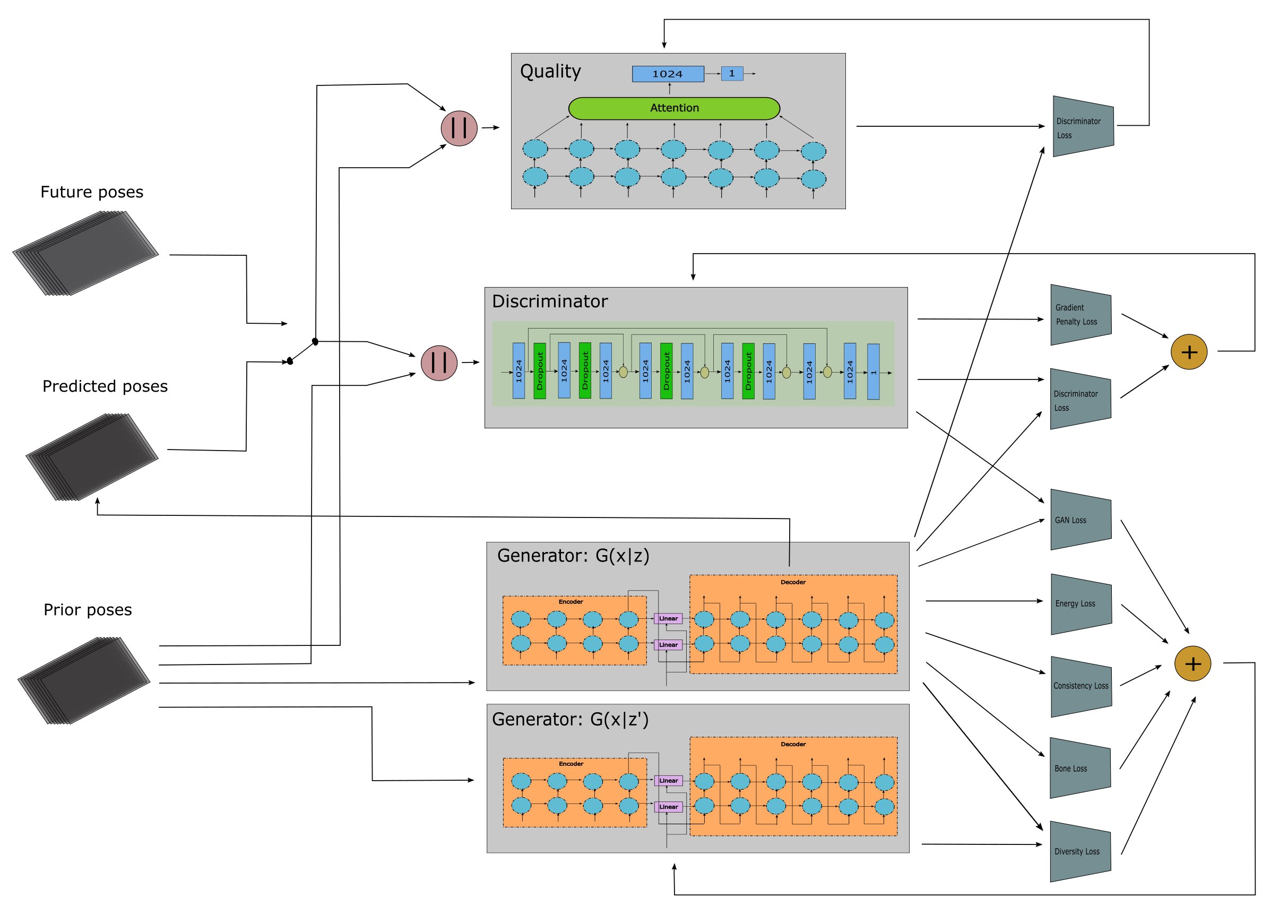}
\caption{Overall architecture of our prediction model. The discriminator is a feed-forward network with skip connections, which takes the concatenation of the input sequence and the predicted sequence, and classifies the concatenation as a real or fake human motion. The two generator networks each take a sequence of human poses as input plus a $z$ vector, and generate a sequence of future human poses, where each $z$ value samples a different future. The second generator is governed by a diversity loss, which controls the variety of the predictions produced. The quality network is a recurrent network with temporal attention, which computes the probability that the concatenation sequence is human.  Note that the quality network does not affect the generator.}
\label{figure:architecture}
\end{figure*}

Our approach is based on GAN architecture as shown in Figure~\ref{figure:architecture}, we have a discriminator that distinguish between valid and not valid human poses and a generator that generate future human poses based on an input sequence and a $z$ vector that sample different output sequence. We have multiple additional losses than the standard GAN in order to stabilize the training and control the predicted human motion. The details will be described next section. 

Furthermore, we test the generalization of the feature learned by the discriminator by using the discriminator network on action recognition task and compare the result with the same network but trained from scratch. We show that by using the discriminator network for action recognition, the network reach better accuracy on fewer number of epochs compared to training from scratch.

The generator is shown in Figure~\ref{figure:architecture}, it is a modified version of the sequence-to-sequence~\cite{nips2014:Ilya} network. It takes as input a sequence of human poses, plus a $z$ vector drawn from a uniform or Gaussian distribution $z \sim p_{z}$. The drawn $z$ value is then mapped to the same space as the output states of the encoder. We then simply add the mapped value of $z$ to the encoder states, and use the result as the initial state of the decoder. We map $z$ to each layer of the encoder, and then feed the last output of the encoder to the first input of the decoder. We use GRU~\cite{corr2014:Chung} for our sequence-to-sequence network, we also tried LSTM~\cite{nc1997:Hochreiter} and did not notice any quality difference.

Let $G$ be the network shown in figure ~\ref{figure:architecture}. We have $y = G(x,z;\theta_{g})$, where $\theta_{g}$ are the network parameters that we need to learn. Each drawn value of $z$ will sample different valid future poses from the given input $x$. Most recent human pose predictions using deep RNNs~\cite{iccv2015:Katerina, cvpr2017:julieta} treat human motion prediction as a regression problem. However, solving human motion prediction using regression suffers from two deficiencies. First, it learns one outcome at a time, and as the predicted sequence length increases, this outcome becomes less probable. Second, using the usual $\ell_{1}$ or $\ell_{2}$ loss creates artifacts~\cite{corr2015:Mathieu}, such as predicting the average of multiple possible outcomes which is not the correct prediction.

We use an adversarial training scheme for three reasons. First, it allows the generation of multiple futures from a single past. Second, it allows the generator to be trained without explicitly using the ground truth of real futures. Third, it implicitly learns the cost function for the prediction based on the data. 

\subsection{Generative adversarial networks}

Generative adversarial networks(GAN) was introduced by~\cite{nips2014:Goodfellow}. It is a unsupervised learning technique inspired by the minimax theorem~\cite{pjm1958:Sion}, in which the generator network and the discriminator network try to outdo each other. The training itself alternates between both networks. In the original paper, the generator learns to generate images close to real images, and the discriminator learns to distinguish between the generated image and the real image from the dataset. In a steady state, the discriminator should predict if an image from the generator network is generated or not with $50\%$ probability.

However, the original GAN algorithm is not stable and is difficult to train, because of its use of Jensen-Shannon (JS) divergence as its loss function. JS can result in zero or infinity due to its ratio between two probabilities that might not overlap initially, and can lead to vanishing gradients in the discriminator network. WGAN~\cite{corr2017:Arjovsky} replaces the JS distance with the Earth Mover Distance (EMD), which is defined and continuous almost everywhere. And according to the author, this mitigates the need to carefully maintain a balance between training the discriminator versus the generator. The discriminator in WGAN does not output a probability, and it does not discriminate between synthetic input and real input, which is why the author renamed the discriminator network to critic network.

Nevertheless, WGAN does not address all the concerns, since the critic still must maintain a Lipschitz constraint. In order to do so, the author clips the weights of the network, which adversely affects the quality of the generator. To address this, the WGAN-GP algorithm~\cite{corr2017:Ishaan} replaces weight clipping in the generator with an added penalty to the loss in the critic, which is based on the computed norm of the gradient with respect to the critic input. Although, we were able to verify these improvements in our domain of human pose prediction and HP-GAN~\cite{cvprw2018:Barsoum} used a modified version of WGAN-GP in their training (in order to avoid converge and stabilize the training), we found out that we can make standard GAN as stable as WGAN-GP, by borrowing the penalty loss from WGAN-GP.

There are multiple reasons on why we prefer GAN instead of WGAN-GP. First, GAN discriminator compute a probability value that indicates if the input is real or fake, WGAN and its family uses critic instead of discriminator that compute a floating point number which cannot be interpreted. Second, the loss value in GAN is positive and decrease when the network converge, in WGAN it is difficult to tell if the network converge or not from the loss alone without looking at the actual generated poses. And third, in our experience with WGAN, after converging if we continue training, it can diverge. We did not see this behavior in GAN. 

\subsection{Human pose prediction}

Figure~\ref{figure:architecture} shows the high level architecture diagram of our proposed GAN network, for Human Pose prediction. Our model is based on GAN~\cite{nips2014:Goodfellow} with gradient penalty from WGAN-GP~\cite{corr2017:Ishaan} and human pose specific losses. From Figure~\ref{figure:architecture}, "Future poses" are the ground truth future poses from the dataset, and "Prior poses" are their corresponding previous poses.

\subsubsection{Discriminator network}

\begin{figure*}[ht]
\centering
\includegraphics[width=\textwidth]{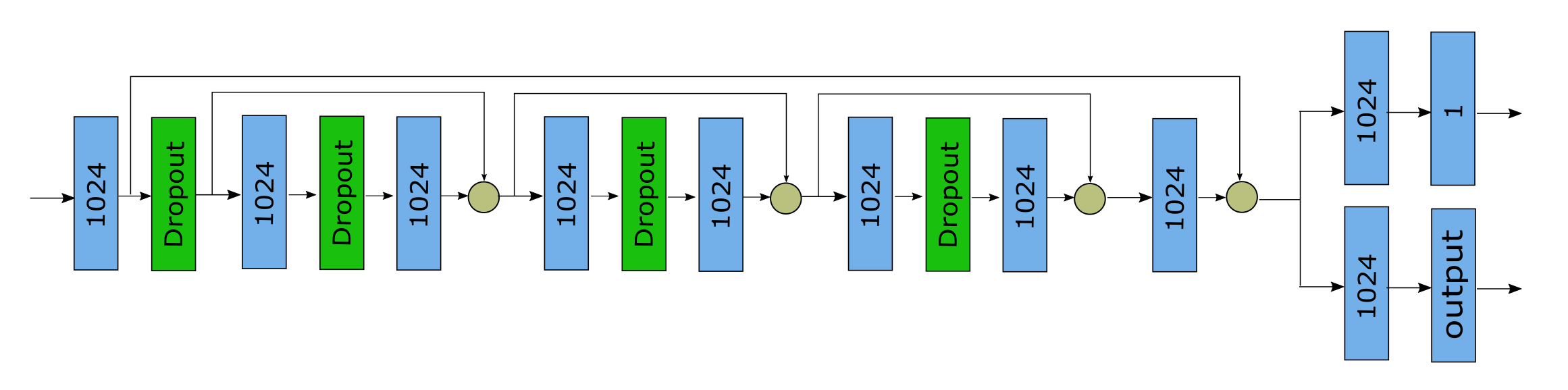}
\caption{Discriminator network is a full connected multi-layer network with a skip connection. It ends with two branches, top branch for GAN training and bottom branch for activity recognition training.}
\label{figure:discriminator_with_branch}
\end{figure*}

The discriminator network, shown in more detail in Figure~\ref{figure:discriminator_with_branch}, is a fully connected feed-forward network with skip connections. It outputs a single probability value between $0$ and $1$, with
$0$ meaning that the input is fake (i.e., generated), and $1$ meaning that the input is real (i.e., taken from the ground truth dataset). 

The input to the discriminator is the concatenation of the prior sequence of human poses, either with the future sequence of human poses from the ground truth, or with the predicted sequence of human poses from the generator network. The job of the discriminator is to compute the probability if a given sequence of human poses is real or fake. The second branch in the discriminator is for action classification; this branch is not used during GAN training.

In addition to its usual role in a GAN, we also use the discriminator network for activity recognition. We show that the features learned by the discriminator during its GAN training help speed up the classification task.

\subsubsection{Generator network}

The Generator is a sequence-to-sequence network as defined in Figure~\ref{figure:architecture}. It takes as input previous human poses and a $z$ vector, and produces a sequence of predicted future human poses. The $z$ vector is a 128-dimensional float vector drawn from a uniform or Gaussian probability distribution.

Prior poses and future poses are concatenated together to form a real pose sequence. Similarly, prior poses and generated poses are concatenated together to form a fake pose sequence. Both real and fake sequences are used for the discriminator GAN loss. GAN loss is used to improve both the discriminator network and the generator network by alternating between discriminator loss and generator loss.

\subsubsection{Quality network}

The quality network, shown in Figure~\ref{figure:architecture}, is a multi-layer recurrent network with temporal attention. The attention layer's purpose is to learn which part of the sequence is more important for generating the probability of the input sequence of poses. Instead of using the more common methods of either averaging the outputs across time, or using the last output in the recurrent network, the attention layer learns from the training data the importance of each output and captures this information in a vector.

\subsection{Human pose prediction: losses}

Prior poses from the ground truth and future poses from the ground truth are concatenated together to form a real pose sequence. Similarly, prior poses from the ground truth and generated poses from the generator are concatenated together to form a fake sequence. Both real and fake sequences are used by the discriminator and the quality network to compute their loses. These and other GAN losses used to improve all three network types, by cycling through discriminator loss, generator loss, and quality loss.

\subsubsection{Discriminator loss}

Based on numerous experiments, almost all GAN algorithms, we tried, were stable by adding gradient penalty from equation~\ref{eq:gradient_penalty} to the discriminator loss similar to the one defined in WGAN-GP. We are not the first to discover that~\cite{corr2017:Fedus}, but we are the first to show it on a complex problem such as skeleton pose prediction. The discriminator loss that we used is as follow:

\begin{equation}
L_{d}=L_{gan}^d+\lambda_{gp} L_{gp}^d+\alpha L_{2}^d
\label{eq:discriminator_total_loss}
\end{equation}

\noindent$L_{gan}^d$ is the original GAN discriminator loss, which is defined as:

\begin{equation}
L_{gan}^d=log(D(x||y))+log(1-D(x||G(x,z)))
\label{eq:discriminator_loss}
\end{equation}
where $||$ indicates concatenation, $x$ is the input sequence, $y$ is the future sequence, and $z$ is a random vector drawn from a uniform distribution.\newline

\noindent$L_{gp}$ is the gradient penalty loss defined as:

\begin{equation}
L_{gp}=(\Vert\nabla_{\hat{x}}D(\hat{x})\Vert_{2} - 1)^{2}
\label{eq:gradient_penalty}
\end{equation}
where $\hat{x}=\epsilon (x||y)+(1-\epsilon)(x||G(x,z))$ and $\epsilon
\sim U[0, 1]$.\newline

\noindent$L_{2}^d$ is the standard $L_{2}^d$ regularization defined as:

\begin{equation}
L_{2}^d=\Vert\theta_{d}\Vert_{2}
\label{eq:critic_l2}
\end{equation}

\noindent In all of our experiments, we set $\lambda_{gp} = 10$ and $\alpha = 0.001$.

\subsubsection{Generator loss}

For the generator loss, we use five components. The first is the standard GAN adversary loss defined in~\cite{nips2014:Goodfellow}. The second is a consistency loss or pose gradient loss, which minimize the delta between two consecutive poses. The third is diversity loss which control the variety between each sampled sequence of poses from the same input but different $z$ value. The fourth is an energy loss in which we try to minimize the energy generated by the motion, this loss is inspired from human motion animation~\cite{pone2015:Nathan}. The fifth is a bone loss, which minimize the delta of bone length between the predicted skeleton and the prior skeleton. We will describe the reason of each of those losses next.

In summary, the generator network total loss is defined as shown below. It is the weighted combination of the five losses.

\begin{equation}
L_{g}=L_{gan}^g+\alpha_{pg} L_{pg}^g+\alpha_{d} L_{d}^g+\alpha_{e} L_{e}^g+\alpha_{b} L_{b}^g
\label{eq:g_total_loss}
\end{equation}

\noindent $L_{gan}^g$ is the generator loss from the original GAN which is defined as:

\begin{equation}
L_{gan}^g=-log(D(x||G(x,z)))
\label{eq:g_loss}
\end{equation}
Where the term $x||G(x,z)$ is the concatenation of the prior poses and the predicted poses from the generator, which then fed to the discriminator.\newline

\noindent $L_{pg}$ is the consistency loss, or pose gradient loss, defined as:

\begin{equation}
L_{pg}^g=\Vert\nabla_{t}y\Vert_{p} = \left[\sum_{t}\vert y_{t}-y_{t-1}\vert^{p}\right]^{1/p}
\label{eq:pose_gradient_loss}
\end{equation}
Equation~\ref{eq:pose_gradient_loss}, $\Vert\nabla_{t}y\Vert_{p}$ computes the gradient over time for the predicted sequence, we use $p=2$ in our training. The reason for this loss is to reduce the delta between skeleton poses between each consecutive frame. For a valid human motion the displacement between each consecutive frame is small.\newline

\noindent $L_{d}^g$ is the diversity loss which is defined as:

\begin{equation}
L_{d}^g=1-\frac{1}{1 + e^-\eta \vert G(x,z_{1})-G(x,z_{2})\vert}
\label{eq:diversity_loss}
\end{equation}
The purpose of the diversity loss, defined in equation~\ref{eq:diversity_loss}, is to control the variation in the predicted poses with the same input of poses but different $z$ value. Therefore, it encourage the trainer to increase the absolute delta $\vert G(x,z_{1})-G(x,z_{2})\vert$ with different $z$ value such as $z_{1} \neq z_{2}$.\newline

\noindent $L_{e}^g$ is the energy loss which is defined as:

\begin{equation}
L_{e}^g=\sum_{t} [\beta_{v}v(t)^2+\beta_{a}a(t)^2]
\label{eq:energy_loss}
\end{equation}
The above equation is based on the energy expenditure defined in~\cite{pone2015:Nathan}, the velocity $v$ and acceleration $a$ are computed from the center of mass in each skeleton pose. The idea here is that for natural human motion the energy expenditure generated by the motion need to be minimum. The same technique used in motion animation. The equation defined in~\cite{pone2015:Nathan} uses velocity and height, which is different than equation~\ref{eq:energy_loss}. We use velocity $v$ and acceleration $a$ is our loss, the reason for that is using height degraded the prediction quality. Because reducing the energy was also reducing the estimated height.\newline

\noindent $L_{b}^g$ is simply the $L_{2}$ norm of the bone length differences between the predicted pose and the ground truth, as shown below.

\begin{equation}
L_{b}^g=\sum_{t}\left[\sum_{i}\vert b^{i}_{t}-b^{i}_{gt}\vert^{2}\right]^{1/2}
\label{eq:bone_loss}
\end{equation}
where $b^{i}_{gt}$ is the ground truth bone length and $b^{i}_{t}$ is the predicted bone length, both at time $t$. We iterate through all bones using index $i$, and sum over all the future skeleton poses using index $t$.\newline

$L_{pg}^g$ is critical during training. If $\alpha_{pg}$ is too large, the effect of $L_{pg}$ loss becomes too high and we obtain less displacement between poses across time; in some cases it can cause copying the same pose. Conversely, we have found that $L_{pg}$ is not critical if we train on a large dataset, like all the 49 classes of NTURGB-D; but for small subset, like one or two classes, then its value becomes critical in avoiding motion discontinuities at the first predicted human pose. Furthermore, to make sure that $L_{pg}^g$ loss does not reach zero, we set a minimum value for the loss as follow $L_{pg}^g=max(C,\Vert\nabla_{t}y\Vert_{p})$, where $C$ is a hyper parameter, the reason is to avoid pushing two consecutive human poses to match exactly. $L_{d}$ is important in order to control the variety between each prediction from the same input sequence but different $z$ value. Without it, we can run into a situation that causes us to obtain identical predictions regardless of the input $z$ value.

\subsubsection{Quality assessment network loss}

We use another network, shown in figure~\ref{figure:architecture}, to judge the quality of the predicted skeleton poses and to decide which model to choose. The quality loss is the same loss used in GAN~\cite{nips2014:Goodfellow}, with the exception that the generator does not use this loss in its training. It is defined as:

\begin{equation}
L_{q}=L_{gan}^q+\alpha L_{2}^q
\label{eq:quality_total_loss}
\end{equation}
where $L_{gan}^q$ is the standard GAN loss defined as:

\begin{equation}
L_{gan}^q=log(Q(x||y))+log(1-Q(x||G(x,z)))
\label{eq:quality_loss}
\end{equation}
and $L_{2}^q$ is the same as $L_{2}^d$ norm used by the discriminator network.

The reason that we use another network to measure the quality of the generator model, instead of using the discriminator network which also output probability that measure if the specified sequence of human poses is real or fake, is that the discriminator is essentially part of the generator loss. Therefore, the generator is optimized to outperform the discriminator, which why we use another network that is not involved in training the generator.

\subsection{Human pose classification}
For the classification part, we used the pre-trained discriminator, from the prediction training, by replacing the last two layers with two new layers for classification, as shown in figure~\ref{figure:discriminator_with_branch}. We train the exact same network twice, one with the pre-trained weights from the discriminator and another with random initialization in order to compare the effect of the unsupervised training on feature learning. The two new layers are fully connected layers, the last one has $k$ outputs, where $k$ is the number of classes used during training.

Since the discriminator learn to differentiate between real and fake human motion, therefore, it learns what constitute a human motion. We postulate that the discriminator learn general features about human motion that can be used on other tasks beside computing the probability that a given sequence of poses are natural or not. In order to cheque this assumption, we train the classifier in figure~\ref{figure:discriminator_with_branch} twice, one with the discriminator weights and another with random initialization, and show that using the discriminator as the basis for the classifier train speed up the training compared to training from scratch.

\subsection{Training}
For the training algorithm, we follow closely most GAN training methods. Inside the training loop, we iterate $k$ times on the discriminator network, and one time on the generator and quality network. We use $k = 10$. We have tried different iteration values, and have tried to dynamically update the iteration count based on the losses of the discriminator and the generator, but none of those methods made any noticeable improvement.

In order to make the quality network training procedure more stable, we reduced its learning rate by half, compared to the discriminator and generator learning rate. For all three networks, we use ADAM~\cite{corr2014:Kingma}, and set the learning rate as $5e-5$ for the discriminator and generator network, and half of that for the quality network. As for the supervised training, we train the network shown in figure~\ref{figure:discriminator_with_branch} twice, one with the same weights as the discriminator and another with random initialization. We use ADAM~\cite{corr2014:Kingma} for the optimizer. The training was done in two phases:
\begin{enumerate}
\item \textbf{Unsupervised phase:} this is the GAN training using the losses described previously, this training train the discriminator, generator and quality network.
\item \textbf{Supervised phase:} this is the action classification training, we update the discriminator network as shown in figure~\ref{figure:discriminator_with_branch}.
\end{enumerate}

\section{Experiments}
To verify our model capability, we run multiple experiments on two of the largest human motion datasets: a Microsoft Kinect dataset NTURGB-D~\cite{cvpr2016:Shahroudy} and a motion capture (MoCap) dataset Human3.6M~\cite{pami2014:Ionescu,iccv2011:Ionescu}. The human poses in NTURGB-D dataset are based on skeleton data from Kinect, which is not perfect due to occlusions, people carrying objects or interacting with other person. However, even with noisy skeletons our model generalizes well on this dataset.

The NTURGB-D action recognition dataset consists of 56,880 actions, and each action comes with the corresponding RGB video, depth map sequence, 3D skeletal data, and infrared video. We use only the 3D skeleton data. They contain the 3D locations of 25 major body joints at each frame, as defined by the Microsoft Kinect API. NTURGB-D has 60 action classes and 40 different subjects, and each action was recorded by three Kinects from different viewpoints. From those 60 classes, 49 classes are for single person actions.

Human3.6M contains 3.6 million 3D human poses and their corresponding images, captured by a Vicon MoCap system. Each of these skeletons has 32 joints. The actions were performed by 11 professional actors covering 17 action classes. Using the code from~\cite{cvpr2017:julieta}, we read the Human3.6m skeleton data and converted it from its angle representation to absolute 3D joint positions.

So for both datasets, we trained our model directly on the absolute 3D joint locations. We fed our model a 3D point cloud, and from this training data the model learns the relationships between the joints in order to predict a valid human pose. This is more difficult than training on the angle, which has less degrees of freedom. We train directly on the joint positions in order to use the same pipeline for both NTU-RGB-D and Human3.6m datasets, and to have a more generic model.

\subsection{Pre-processing}

For preprocessing, we computed the mean and standard deviation across the entire dataset, then substract each joint with mean and divide by double the standard deviation. The reason for dividing by double the standard deviation, is to reduce most join range to be within [-1, 1]. We tried other normalization technique such as normalizing each joint to the range of [-1, 1] by using the dimensions of the Kinect frustum at its maximum 5 meter depth for NTU-RGBD dataset. Or normalize the join using min and max values.

The Human3.6M dataset has fewer clips than the NTURGB-D dataset, however each clip is much longer. In order to use the same pipeline for both datasets, we split Human3.6M clips into shorter segments and only use every other frame in our training. During training, the segment is chosen at random.

We use 300 epochs for the NTU-RGBD dataset and 1000 epochs for the Human3.6M dataset due to its small number of clips. For Human3.6M dataset we select a random segment during each iteration, therefore, each epoch get different set of frames.

\subsection{Quantifying the results}

One of the main problems of GANs is that the loss does not provide any indication of the quality of the generated data. According to the authors of WGAN~\cite{corr2017:Arjovsky} and WGAN-GP~\cite{corr2017:Ishaan}, one of the improvements that WGAN made on the original GAN is that their loss value does in fact provide a quality measure. In our human motion prediction problem, the loss provides some indication of how much the generated sequence looks like a valid human pose. However, we have observed that this is not strictly monotonic; a smaller loss does not always indicate a better quality in WGAN or WGAN-GP. Even worth, once the model reach a good convergence state, further training will cause the model to diverge. Using GAN with gradient penalty help mitigate this problem, with it the loss value indicate the quality if the generator and with GAN the discriminator estimate probability which can be used to measure the quality of the estimated human motion.  

Nonetheless, the reason that we chose to use another network, to measure the quality of the predicted human poses, is because the discriminator act as a loss to the generator. Which mean the generator is optimized to reduce the accuracy of the discriminator. This is why we added another network for the whole purpose to measure the quality if the predicted skeleton poses, we call it quality network. The loss of the generator does not use the quality network. Therefore, we added a quality network, whose sole purpose is to learn the probability that a given sequence is a valid human motion. To find the best model, inside the training loop we generate $N$ predictions and compute the probability, by evaluating the quality network on each of those predictions. Then, we compute the number of predictions $k$ that has probability more than 50\%. We keep track of the model with maximum $k$ during training, we only start tracking the best model after certain number of epochs, in order to give the network a chance to learn evaluating human motion.

\section{Results}

\begin{figure*}[ht]
  \centering
  \begin{subfigure}[t]{\textwidth}
    \raisebox{-\height}{\includegraphics[width=\textwidth]{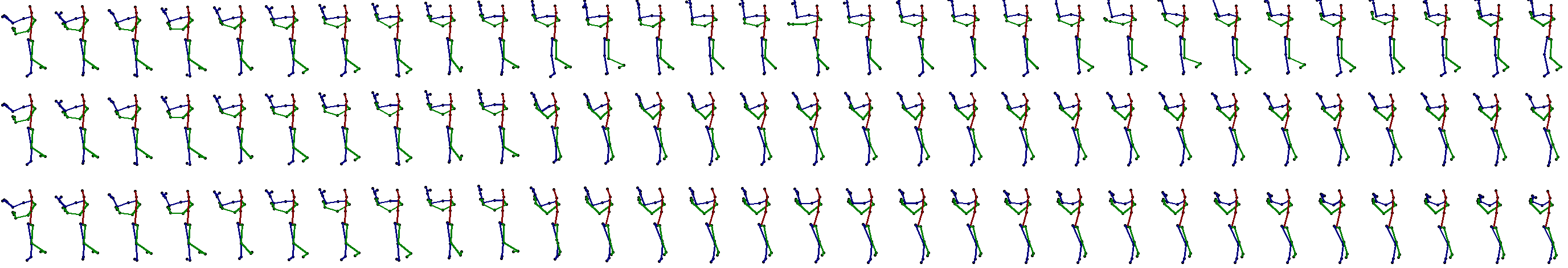}}
    \caption{Sample prediction from NTURGB-D dataset.}
  \end{subfigure}
  \hfill
  \begin{subfigure}[t]{\textwidth}
    \raisebox{-\height}{\includegraphics[width=\textwidth]{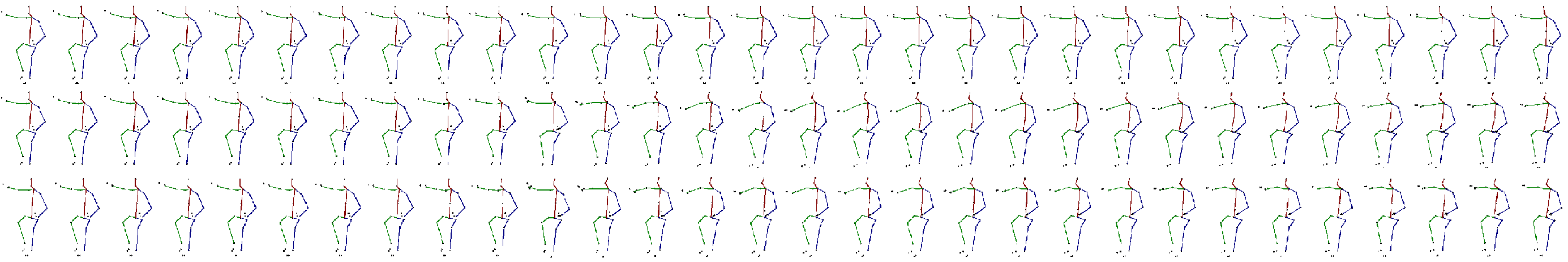}}
    \caption{Sample prediction from Human3.6M dataset.}
  \end{subfigure}
\caption{Example of poses prediction from the generator, input 10 poses and output 20 poses. First row is the ground truth and the other two are predicted poses per $z$ value.}
\label{figure:pred}
\end{figure*}

Figure~\ref{figure:pred} shows the prediction results for both NTURGB-D and Human3.6M datasets. The top row is the ground truth, and each subsequent row corresponds to the predicted human poses from different $z$ value drawn from a uniform distribution. The input sequence of poses is of size 10 and the generator predict 20 output poses in this example. As shown in the figure, each $z$ value generates a separate possible future sequence of human poses. The first few of the predicted poses are very close to the ground truth, which is expected. As we predict more poses in the future, they start to diverge from the ground truth. 

\begin{figure}[ht]
  \centering
  \begin{subfigure}[b]{0.49\linewidth}
    \includegraphics[width=\textwidth]{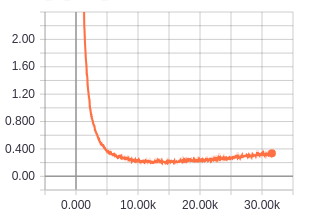}
    \caption{Discriminator loss}
    \label{figure:discriminator_loss}
  \end{subfigure}
  \begin{subfigure}[b]{0.49\linewidth}
    \includegraphics[width=\textwidth]{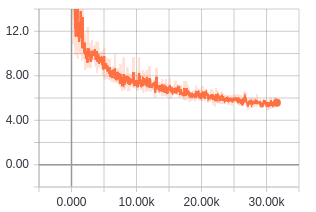}
    \caption{Generator loss}
    \label{figure:generator_loss}
  \end{subfigure}
    
  \begin{subfigure}[b]{0.49\linewidth}
    \includegraphics[width=\textwidth]{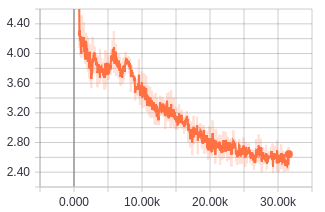}
    \caption{Quality loss}
    \label{figure:quality_loss}
  \end{subfigure}
  \begin{subfigure}[b]{0.49\linewidth}
    \begin{subfigure}[b]{0.49\textwidth}
      \includegraphics[width=\textwidth]{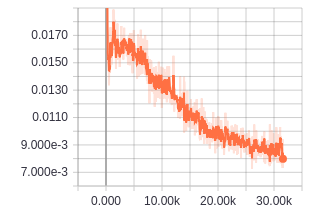}
      \caption{Consistency}
      \label{figure:consistency_loss}
    \end{subfigure}
    \begin{subfigure}[b]{0.49\textwidth}
      \includegraphics[width=\textwidth]{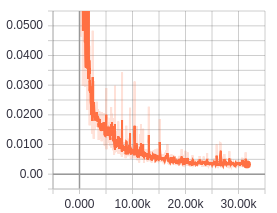}
      \caption{Energy}
      \label{figure:energy_loss}
    \end{subfigure}
    \begin{subfigure}[b]{0.49\textwidth}
      \includegraphics[width=\textwidth]{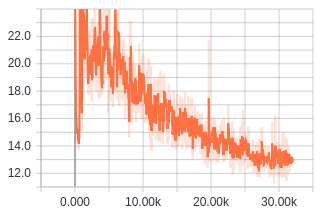}
      \caption{Bone}
      \label{figure:bone_loss}
    \end{subfigure}
    \begin{subfigure}[b]{0.49\textwidth}
      \includegraphics[width=\textwidth]{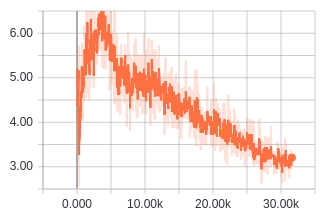}
      \caption{GAN}
      \label{figure:gan_loss}
    \end{subfigure}      
  \end{subfigure}
  \caption{The above is the plot of various losses during training, the $x$ axis corresponding to the step.}
  \label{figure:loss_plot}
\end{figure}

In figure~\ref{figure:loss_plot}, we plot the loss values as a function of the batch axis. Compared to the work done by~\cite{cvprw2018:Barsoum} using WGAN-GP, our loss converge and the convergence map to the quality of the prediction. The combination of GAN and gradient penalty enabled us to achieve that.

\begin{figure}[htb]
\centering
\includegraphics[width=0.75\linewidth]{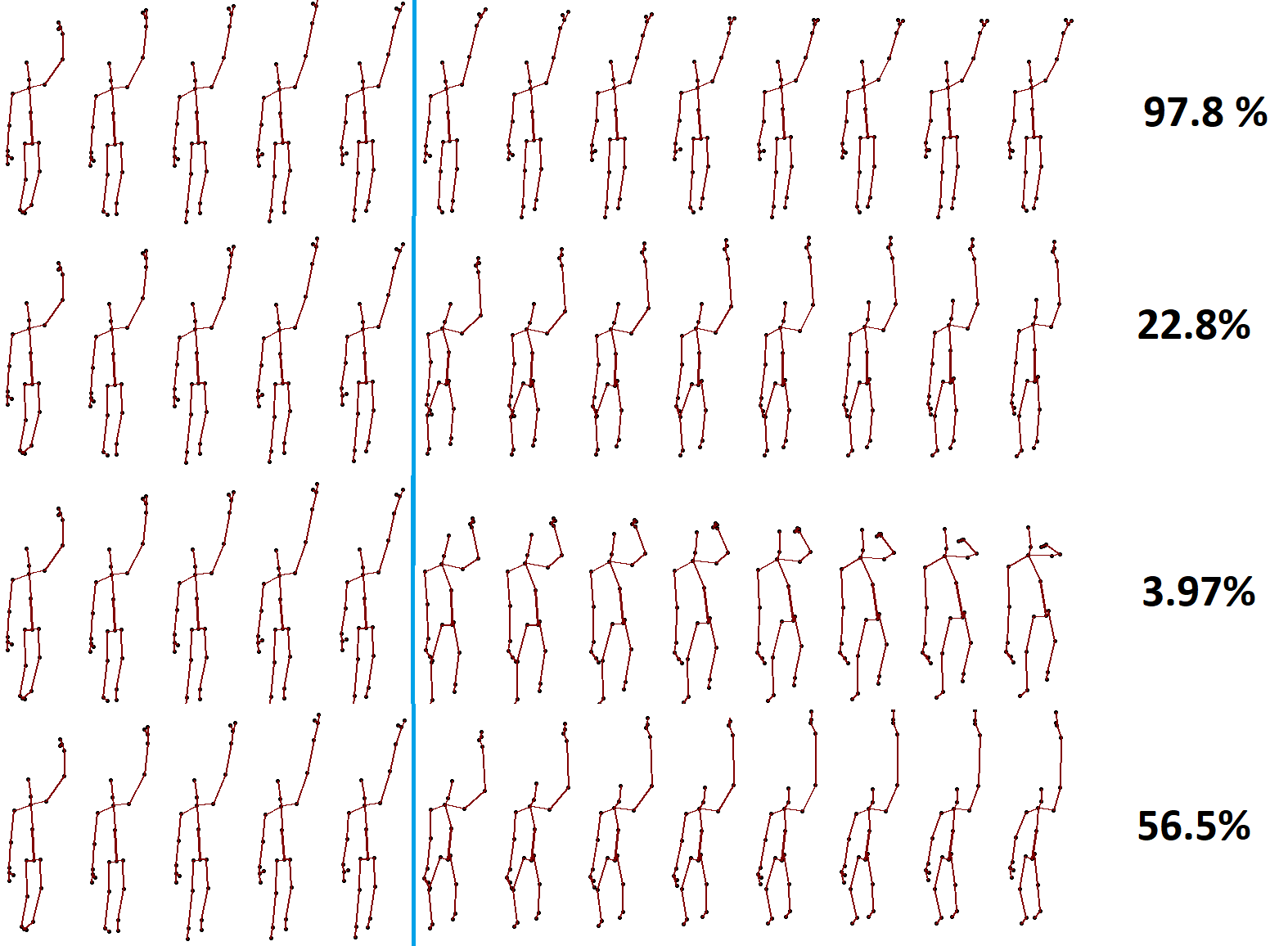}
\caption{The probability that a sequence generated from the first five poses is a valid human motion.}
\label{figure:ntu_pred_prob}
\end{figure}

From figure~\ref{figure:quality_loss}, we can see that the quality loss also converge. It therefore can be used to measure the quality of the prediction, as shown with the four sequences in Figure~\ref{figure:ntu_pred_prob}. As the visual quality of the prediction worsens, the computed probability of a valid human pose also approaches zero. The accuracy of this estimation enabled us to select the best model from the training.

As an example, the third row in the figure, with low quality probability $3.97\%$, generates bone lengths that are visually different than the ground truth, and the motion looks unnatural. We can easily see that the other rows provide better human poses, and that the higher the value, the better the quality of the poses across the time axis. For example, one difference between the first and second rows is that the transition from the ground truth in the first row is better, as shown in the poses before and after the blue line.

More detailed assessments of the quality network are shown in Table~\ref{table:quality_output}, which shows the probability output of the quality network for the ground truth pose and $7$ different predictions (under a different $z$ value per column), at various epochs (per row).

\begin{table}[h!]
\centering
  \begin{tabular}{|c|c|c|c|c|c|c|c|c|}
  \hline
  \multirow{2}{*}{Epoch} & \multirow{ 2}{*}{Real} & \multicolumn{7}{c|}{Human pose predictions for different $z$ values} \\\cline{3-9}
  & & 1 & 2 & 3 & 4 & 5 & 6 & 7 \\
  \hline
  1 & 0.48 & 0.51 & 0.51 & 0.51 & 0.51 & 0.51 & 0.51 & 0.51 \\
  \hline
  2 & 0.71 & 0.57 & 0.57 & 0.57 & 0.57 & 0.57 & 0.57 & 0.57 \\
  \hline
  13 & 0.96 & 0.06 & 0.06 & 0.07 & 0.05 & 0.06 & 0.06 & 0.06 \\
  \hline
  14 & 0.95 & 0.07 & 0.06 & 0.08 & 0.05 & 0.07 & 0.06 & 0.06 \\
  \hline
  152 & 0.88 & 0.59 & 0.31 & 0.23 & 0.65 & 0.21 & 0.51 & 0.18 \\
  \hline  
  \end{tabular}
  \caption{The probability output of the quality assessment network after each epoch (per row) for the ground truth (Real) and $10$ predictions (per column).}
  \label{table:quality_output}
\end{table}

At Epoch 1, the quality network can not differentiate between real and non-real human poses, so the output is like flipping a coin with probability $p_{i}\approx 50\%$.  But neither can the generator generate valid sequences of poses.  At Epoch 13 and 14, the quality network quickly learns to discard the generated poses, and so we have the probability of the ground truth is high $\approx 95\%$ and the probability of the prediction is close to zero.  At Epoch 152, the generator has learned to generate non-real human poses that are close to the real ones, so we have the probability of the predicted values is increasing, and some of the predicted poses exceed $50\%$.

\subsection{Classification}

\begin{figure}
\centering
\includegraphics[width=\linewidth]{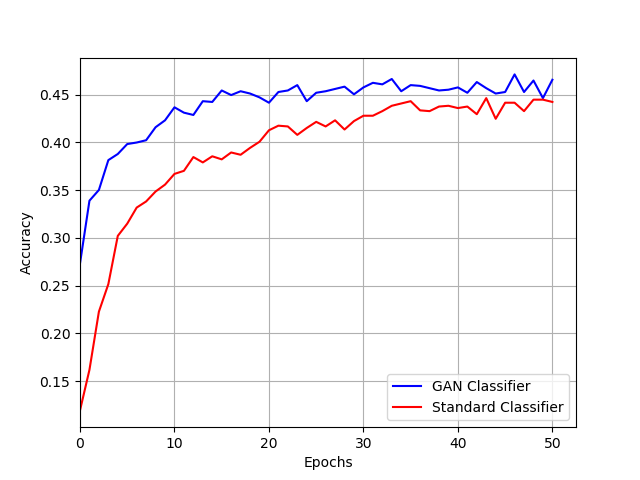}
\caption{The blue plot is the classifier initialized with the discriminator weights, and the red plot is the classifier initialized at random. As shown, the one using the discriminator parameters start better and reach steady state much faster.}
\label{figure:supervised_accuracy}
\end{figure}

To verify that the features learned by the discriminator is generic human motion feature and can be applied to other related tasks. We train a classifier using the discriminator network with the last two layer replaced as shown in figure~\ref{figure:discriminator_with_branch}. We train the discriminator twice, one using the parameters learned from the unsupervised training and another with the parameters initialized at random. As shown in figure~\ref{figure:supervised_accuracy}, we train on the $49$ classes of the NTURGB-D dataset. The blue plot, which uses the discriminator weights, start at around $27.5\%$ accuracy and reach steady state at around epoch $20$. The red plot which was trained from scratch, start at much lower accuracy and took double the number of epochs to reach steady state.

\subsubsection{Unseen classes}

\begin{figure}[htb]
  \centering
  \begin{subfigure}[b]{0.49\linewidth}
    \includegraphics[width=\textwidth]{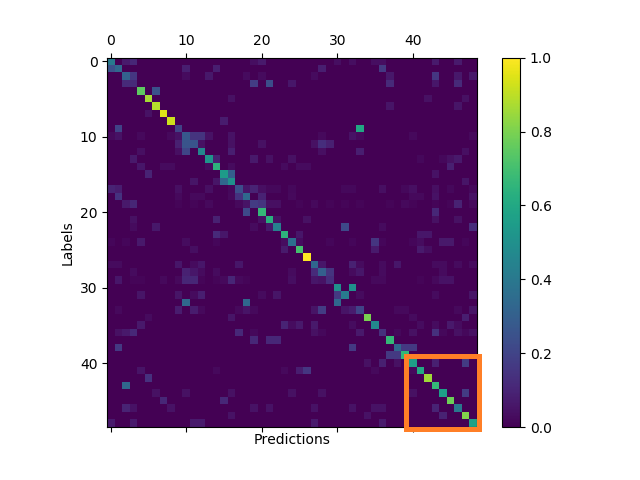}
  \end{subfigure}
  \begin{subfigure}[b]{0.49\linewidth}
    \includegraphics[width=\textwidth]{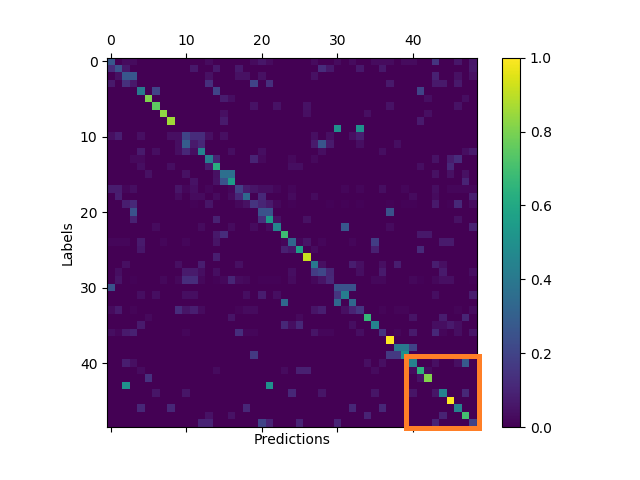}
  \end{subfigure}
\caption{Human3.6m classifier. Comparison between the confusion matrix at epoch 12, with some classes not used in the GAN training. Left side is using the pre-trained discriminator and right side is training from scratch.}
  \label{figure:act_cm_discriminator_12}
\end{figure}

To verify that the learned features are generic motion figures, valid even for action categories not seen during training, we removed the last $10$ classes from the NTURGB-D dataset during GAN training, and restored them only in the classification training. As shown in Figure~\ref{figure:act_cm_discriminator_12}, the left confusion matrix was trained with the discriminator network and the right was trained from scratch. The orange rectangles outline classes $40$ to $49$, which were the labels not used during the GAN training. Even within these rectangles, it is easy to see that the left confusion matrix has better accuracy and fewer off-diagonal terms than the corresponding area in the right confusion matrix. 

The maximum accuracy on the test data using the pre-trained weight of the discriminator is $49.20\%$ at epoch $276$, and the maximum accuracy training from scratch using the same network architecture is $48.64\%$ at epoch $339$.

\subsubsection{Effect of reducing training data size}

One of the advantages of using a pre-trained model on a different task is that the amount of labels needed for that task are smaller than training from scratch. We tested how good were the features learned by the discriminator as a result of GAN training on a reduced training data size. We reduced the training data by $50\%$ and by $25\%$, then reran our experiments.

\begin{figure}[htb]
  \centering
  \begin{subfigure}[b]{0.49\linewidth}
    \includegraphics[width=\textwidth]{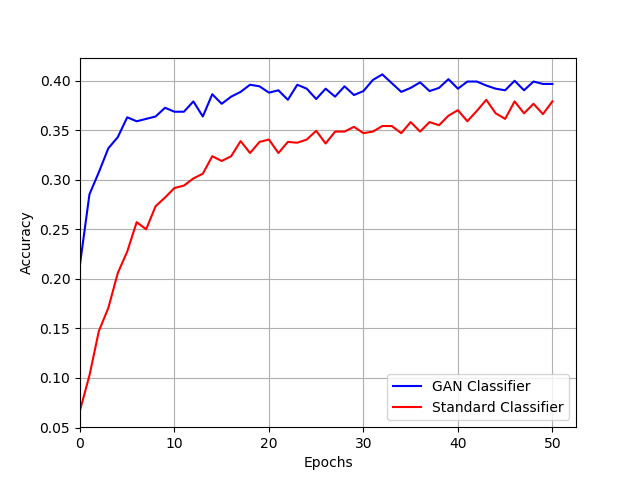}
  \end{subfigure}
  \begin{subfigure}[b]{0.49\linewidth}
    \includegraphics[width=\textwidth]{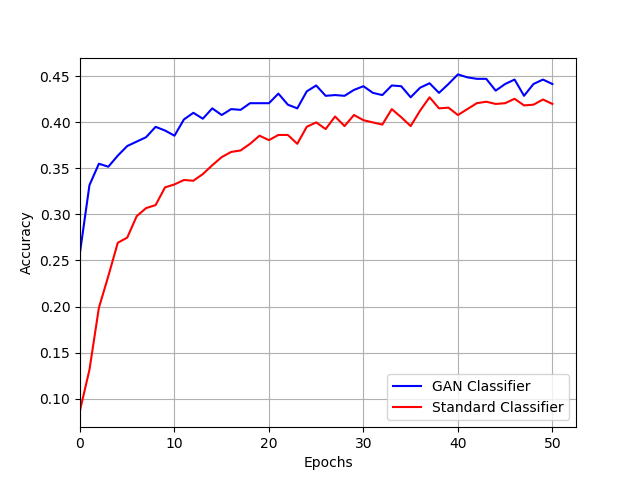}
  \end{subfigure}
\caption{NTURGB-D accuracy with reduced training data. Left side, we reduced the training data by 50\%, and right side, we reduced the training data by 25\%, in order to show the effect of the pre-training on reduced labeled data.}
  \label{figure:ntu_nn_accuracy_with_smaller_training_data}
\end{figure}

As shown in Figure~\ref{figure:ntu_nn_accuracy_with_smaller_training_data}, when we reduced the training size by $50\%$, both training from the discriminator and training from scratch became less accurate. But the discriminator network still performed better, and reached $40\%$ accuracy. Reducing the training data by $25\%$, the discriminator reached $46.71\%$ accuracy, which is not far from training on the full dataset. Training from scratch reached only $42.07\%$, showing that the amount of training data needed using the pre-trained discriminator is less than training from scratch, while
reaching the same accuracy.

\subsection{Ablation Study}

\begin{figure}
  \centering
  \begin{subfigure}[t]{\linewidth}
    \includegraphics[width=\linewidth]{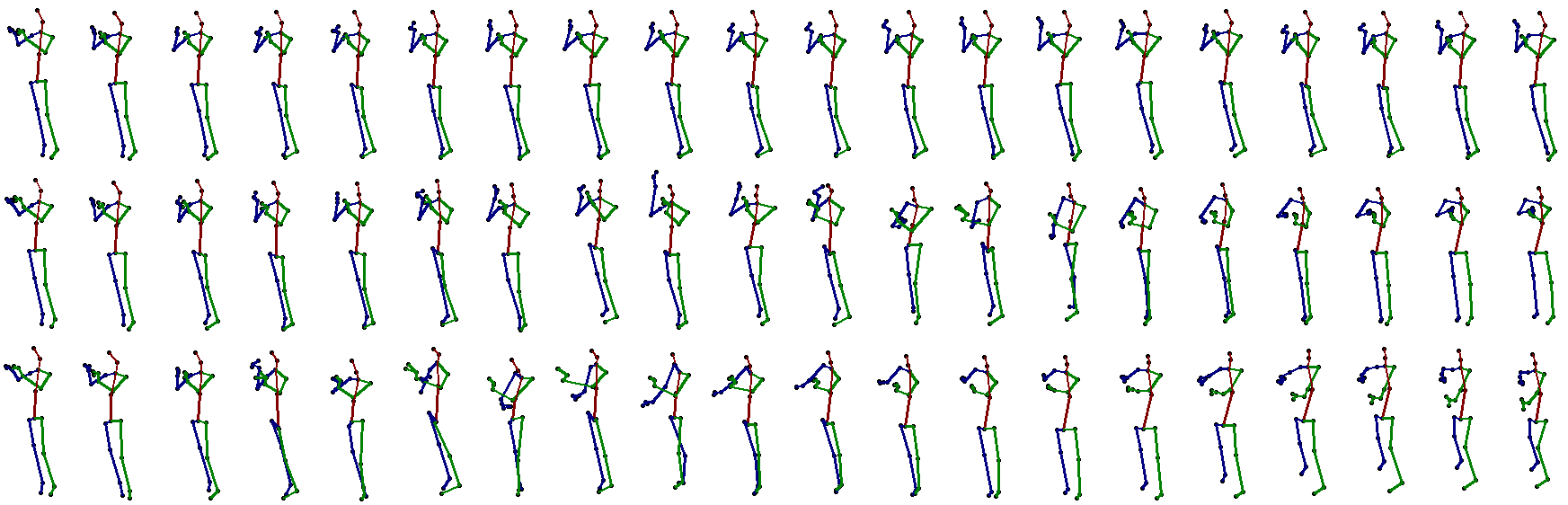}
    \caption{Pose generated with divergence loss.}
  \end{subfigure}
  \hfill
  \begin{subfigure}[t]{\linewidth}
    \includegraphics[width=\linewidth]{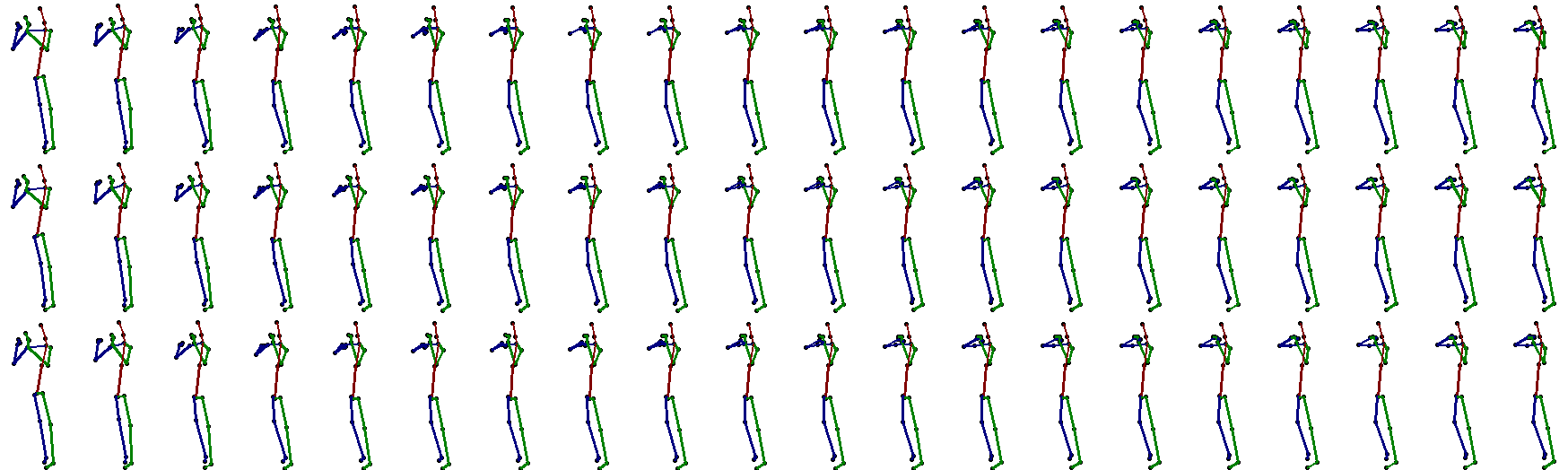}
    \caption{Pose generated without divergence loss.}
  \end{subfigure}
\caption{A comparison on the effect of the divergence loss on the prediction result. Each row is from different $z$ value.}
\label{figure:pred_divergence_ablation}
\end{figure}

\noindent\textbf{Divergence loss:} divergence loss control the diversity of the predicted poses between $z$ values from the same input sequence. Increasing the coefficient of the divergence loss, will cause each $z$ value produce different output sequence, removing the divergence loss will cause each $z$ value produce almost the same sequence of human poses as shown in figure~\ref{figure:pred_divergence_ablation}.

\noindent\textbf{Consistency loss:} this loss control the pace of the motion between frames. Without it we notice jumpiness in the predicated poses in some cases.

\noindent\textbf{Gradient penalty:} in GAN training, without the gradient penalty, we could not get the model to converge and it collapses easy. We tried other type of GAN algorithm that are not based on earth mover distance and all of them performed better with the gradient penalty.

\noindent\textbf{GAN versus WGAN-GP:} gradient penalty was introduced in the improved WGAN, both GAN and WGAN-GP produce similar quality. But based on multiple experiments on NTURGB-D and Human3.6M dataset, the loss of WGAN-GP did not indicate the quality of the generator and even after reaching a good prediction quality it diverge if we continue training.

\section{Conclusions and future work}

We have shown a novel sequence-to-sequence model for human motion prediction, with the ability to control the prediction behavior, in addition to predict multiple plausible future of human poses from the same input. We also showed the representation power learned by the discriminator which can be used in other related tasks such as action classification. To quantify the quality of the non-deterministic predictions, we simultaneously trained a motion-quality-assessment model that learns the probability that a given skeleton sequence is a real human motion. We tested our architecture on two different datasets, one based on the Kinect sensor and the other based on MoCap data. Experiments show that our model performs well on both datasets.

Building on this work, we plan to investigate the semantic meaning and space of the $z$ vector. If we can compute the reverse mapping from pose sequence to $z$, we might be able to use $z$ values for action classification or clustering. Another area of exploration is to use the generated data for augmentation during training in order to increase pose variety.

\clearpage


\bibliographystyle{spmpsci}
\bibliography{references}

\end{document}